\pdfoutput=1

\documentclass[11pt]{article}

\usepackage[]{EMNLP2023}

\usepackage{times}
\usepackage{latexsym}

\usepackage[T1]{fontenc}

\usepackage[utf8]{inputenc}

\usepackage{microtype}

\usepackage{inconsolata}
\usepackage{graphicx}
\usepackage{multirow}
\usepackage{arabtex}
\usepackage{utf8}
\setcode{utf8}
\usepackage{multirow,tabularx}
\graphicspath{ {./figures/} }
\usepackage{changepage}

\usepackage{lipsum}
\usepackage{placeins}
\usepackage{float}

\usepackage{makecell}

\usepackage{multirow}
\usepackage{amsmath}
\usepackage{amsfonts}
\usepackage{amssymb}
\usepackage{graphicx}
\usepackage{wrapfig}
\graphicspath{ {./figures/} }

\usepackage{abstract}
\usepackage{caption}
\usepackage{subcaption}
\usepackage{array}
\usepackage{xspace}

\newcommand{\Ar}[1]{{\small \<#1>\xspace}}

\newcommand{\wojoodf}{\cal{\textit{Wojood$_{Fine}$}}\xspace}

\newcommand{\tagg}[1]{{\small #1\xspace}}

\usepackage[colorinlistoftodos,prependcaption,textsize=tiny]{todonotes}

\title{Arabic Fine-Grained Entity Recognition}

\author{Haneen Abdallatif Liqreina \\
  Birzeit University \\ Birzeit, Palestine \\
  \texttt{1195325@student.birzeit.edu} \And
  Mustafa Jarrar \\
  Birzeit University \\ Birzeit, Palestine \\
  \texttt{mjarrar@birzeit.edu} \And
Mohammed Khalilia \\
  Birzeit University \\ Birzeit, Palestine \\
  \texttt{ mkhalilia@birzeit.edu} \AND 
Ahmed Oumar El-Shangiti \\
   MBZUAI   \\  Abu Dhabi, United Arab Emirates \\ \texttt{ahmed.oumar@mbzuai.ac.ae} \And
Muhammad Abdul-Mageed  \\
  UBC and MBZUAI \\ Vancouver, Canada \\
  \texttt{muhammad.mageed@ubc.ca} }

\begin{document}
\maketitle
\begin{abstract}
\noindent Traditional NER systems are typically trained to recognize coarse-grained entities, and less attention is given to classifying entities into a hierarchy of fine-grained lower-level subtypes. This article aims to advance Arabic NER with fine-grained entities. We chose to extend Wojood (an open-source Nested Arabic Named Entity Corpus) with subtypes. In particular, four main entity types in Wojood, geopolitical entity \tagg{(GPE)}, location \tagg{(LOC)}, organization \tagg{(ORG)}, and facility \tagg{(FAC)}, are extended with $31$ subtypes. To do this, we first revised Wojood's annotations of \tagg{GPE}, \tagg{LOC}, \tagg{ORG}, and \tagg{FAC} to be compatible with the LDC's ACE guidelines, which yielded $5,614$ changes. Second, all mentions of \tagg{GPE, LOC, ORG,} and \tagg{FAC} ($\sim$ $44K$) in  Wojood are manually annotated with the LDC's ACE subtypes. We refer to this extended version of Wojood as \wojoodf. To evaluate our annotations, we measured the inter-annotator agreement (IAA) using both Cohen's Kappa and $F_1$ score, resulting in $0.9861$ and $0.9889$, respectively. To compute the baselines of \wojoodf, we fine-tune three pre-trained Arabic BERT encoders in three settings: flat NER, nested NER and nested NER with subtypes and achieved $F_1$ score of $0.920$, $0.866$, and $0.885$, respectively. Our corpus and models are open-source and available at {\scriptsize  \url{https://sina.birzeit.edu/wojood/}}.
\end{abstract}

\section{Introduction}
\label{sec:intro}

Named Entity Recognition (NER) is the task of identifying and classifying named entities in unstructured text into predefined categories such as people, organizations, locations, disease names, drug mentions, among others  \citep{li-jing20}.
NER is widely used in various applications such as information extraction and retrieval \citep{jiang-etal-2016-evaluating}, question answering \citep{liu2020asking}, word sense disambiguation \cite{JMHK23,HJ21b}, machine translation \citep{jain2019entity,nlp23}, automatic summarization \citep{summerscales2011automatic,nlp23},  interoperability \cite{JDF11} and cybersecurity \citep{10.1007/978-3-030-51310-8_2}.

Traditional NER systems are typically trained to recognize coarse and high-level categories of entities, such as person (\tagg{PERS}), location (\tagg{LOC}), geopolitical entity (\tagg{GPE}), or organization (\tagg{ORG}). However, less attention is given to classifying entities into a hierarchy of fine-grained lower-level subtypes \cite{zhu2020fine, fg-ner}. For example, locations (\tagg{LOC}) like Asia and Red Sea could be further classified into \tagg{Continent} and \tagg{Water-Body}, respectively. Similarly, organizations like Amazon, Cairo University, and Sphinx Cure can be classified into \tagg{commercial}, \tagg{educational}, and \tagg{health} entities, respectively. Belgium, Beirut, and Brooklyn can be classified into \tagg{Country}, \tagg{Town}, and \tagg{Neighborhood} instead of classifying them all as \tagg{GPE}. The importance of classifying named entities into subtypes is increasing in many application areas, especially in question answering, relation extraction, and ontology learning \citep{10.1007/11880592_49}.

As will be discussed in the following sub-section, the number of NER datasets that support subtypes is limited, particularly for the Arabic language. The only available Arabic NER corpus with subtypes is the LDC's ACE2005 \cite{ACE2005}. However, this corpus is expensive. In addition, ACE2005 was collected two decades ago and hence may not be representative of the current state of Arabic language use. This is especially the case since language models are known to be sensitive to temporal and domain shifts (see section \ref{sec:analysis}).

To avoid starting from scratch, we chose to extend upon a previously published and open-source Arabic NER corpus known as 'Wojood' \cite{JKG22}. Wojood consists of $550K$ tokens manually annotated with $21$ entity types.  In particular, we manually classify four main entity types in Wojood (\tagg{GPE}, \tagg{LOC}, \tagg{ORG}, and \tagg{FAC}) with $31$ new fine-grained subtypes. This extension is not straightforward as we have to change ($5,614$ changes) the original annotation of these four types of entities to align with LDC guidelines before extending them with subtypes. The total number of tokens that are annotated with the $31$ subtypes is $47.6$K. Our extended version of Wojood is hereafter called \wojoodf. We measure inter-annotator agreement (IAA) using both Cohen’s Kappa and $F_1$, resulting in $0.9861$ and $0.9889$, respectively. 

To compute the baselines for \wojoodf, we fine-tune three pre-trained Arabic BERT encoders across three settings: (i) flat, (ii) nested without subtypes, and (iii) nested with subtypes, using multi-task learning. Our models achieve $0.920$, $0.866$, and $0.885$ in $F_1$, respectively.

The remaining of the paper is organized as follows: Section \ref{sec:related_work} overviews related work, and Section \ref{sec:corpus} presents the \wojoodf corpus, the annotation process, and the inter-annotator-agreement measures. In Section \ref{sec:modelingNER}, we present the experiments and the fine-tuned NER models. In Section \ref{sec:analysis} we present error analysis and out-of-domain performance and we conclude in Section \ref{sec:conclusion}.

\section{Related Work}
\label{sec:related_work}

Most of the NER research is focused on coarse-grained named entities and typically targets a limited number of categories. For example, \citet{chinchor1997muc} proposed three classes: person, location and organization. The Miscellaneous class was added to CoNLL-2003 \cite{sang2003introduction}. Additional four classes (geo-political entities, weapons, vehicles, and facilities) were also introduced in the ACE project \cite{ACE2005}. The OntoNotes corpus is more expressive as it covers $18$ types of entities \cite{ontonotes5}.

Coarse-grained NER is a good starting point for named entity recognition, but it is not sufficient for tasks that require a more detailed understanding of named entities \cite{ling2012fine,hamdi2021multilingual}.

Substantial research has been undertaken to identify historical entities. 
For instance, the HIPE shared task \cite{ehrmann2020extended} focused on extracting named entities from historical newspapers written in French, German, and English.
One of its subtasks was the recognition and classification of mentions according to finer-grained entity types.
The corpus used in the shared task consists of tokens annotated with five main entity types and $12$ subtypes, following the IMPRESSO guidelines \cite{ehrmann2020impresso}. 
A similar corpus, called NewsEye, was collected from historical newspapers in four languages: French, German, Finnish, and Swedish \cite{hamdi2021multilingual}. 
The corpus is annotated with four main types: \tagg{PER}, \tagg{LOC}, \tagg{ORG}, and \tagg{PROD}. The \tagg{LOC} entities were further classified into five subtypes, and the \tagg{ORG} entities into two subtypes. \citet{fg-ner} proposed a one million fine-grained NER corpus for Dutch, which was annotated using six main entity types and $27$ subtypes ($10$ subtypes for \tagg{PERS}, three for \tagg{ORG}, nine for \tagg{LOC}, three for \tagg{PROD}, and two for events).  

\citet{zhu2020fine} noted that NER models cannot effectively process fine-grained labels with more than $100$ types. Thus, instead of having many fine-grained entities at the top level, they propose a tagging strategy in which they use $15$ main entity types and $131$ subtypes. Additionally, \citet{ling2012fine} proposed a fine-grained set of $112$ tags and formulated the tagging problem as multi-class multi-label classification. 

A recent shared task was organized by \citet{fetahu2023semeval} at SemEval-2023 Task 2, called MultiCoNER 2 (Fine-grained Multilingual Named Entity Recognition). A multilingual corpus (MULTICONER V2) was extracted from localized versions of Wikipedia covering $12$ languages - Arabic is not included. The corpus was annotated with a NER taxonomy consisting of $6$ coarse-grained types and $33$ fine-grained subtypes (seven subtypes for Person, seven for Group, five for PROD, five for Creative Work, and five for Medical). Most participating systems outperformed the baselines by about $35\%$ $F_1$.

There are a few Arabic NER corpora \cite{DH21}, but all of them are coarse-grained. The ANERCorp corpus covers four entity types \cite{Benajiba2007}, CANERCorpus covers $14$ religion-specific types \cite{Salah2018}, and Ontonotes covers $18$  entities \cite{ontonotes5}. The multilingual ACE2005 corpus \cite{ACE2005}, which includes Arabic, covers five coarse-grained entities and $35$ fine-grained subtypes (3 subtypes for \tagg{PERS}, $11$ for \tagg{GPE}, seven for \tagg{LOC}, nine for \tagg{ORG}, and five for \tagg{FAC}). Nevertheless, the ACE2005 corpus is costly and covers only one domain (media articles) that was collected $20$ years ago. The most recent Arabic NER corpus is Wojood \cite{JKG22}, which covers $21$ nested entity types covering multiple domains. However, Wojood is a coarse-grained corpus and does not support entity subtypes. 

To build on previous research on Arabic NER, we chose to extend the Wojood corpus with finer-grained subtypes. To ensure that our Wojood extension is compatible with other corpora, we chose to follow the ACE annotation guidelines.

\renewcommand\thesection{\arabic{section}}

\section{\wojoodf Corpus}
\label{sec:corpus}

\wojoodf expands the annotation of the Wojood corpus \citep{JKG22}, by adding fine-grain annotations for named-entity subtypes. Wojood is a NER corpus with $550$K tokens annotated manually using $21$ entity types. About $80$\% of Wojood was collected from MSA articles, while the $12$\% was collected from social media in Palestinian and Lebanese dialects (Curras and Baladi corpora \cite{EJHZ22,JHRAZ17,JHAZ14}). One novelty of Wojood is its nested named entities, but some entity types can be ambiguous, which will affect downstream tasks such as information retrieval. For instance, the entity type ``Organization" may refer to the government, educational institution, or a hospital to name a few. That is why \wojoodf adds subtypes to four entity types: Geopolitical Entity  \tagg{(GPE)}, Organization  \tagg{(ORG)}, Location  \tagg{(LOC)}, and Facility  \tagg{(FAC)}. Table \ref{tab:tab2} shows the overall counts of the main four entity types in Wojood and \wojoodf. Note that creating \wojoodf was not a straightforward process as it required revision of the Wojood annotation guidelines, which we discuss later in this section.   
As discussed in \cite{JKG22}, Wojood is available as a RESTful web service, the data and the source-code are also made publicly available \cite{JA19,GJJB23,JAM19,ADJ19,APJ16}.

\begin{table}[h]
\small
\begin{tabular}{p{1.5cm}|p{2.5cm}|p{2.5cm}}
\hline
 \textbf{Tag} & \textbf{Wojood} & \textbf{\wojoodf}   \\ 
  \hline
 
 GPE  &  21,780   &  23,085  
\\ 
  
ORG  &  18,785  &  18,747  
\\ 

LOC  &  917 &  1,441  
\\ 
FAC  &  1,215 &  1,121   
\\ 
 \hline

 \textbf{Total} & \textbf{42,697} & \textbf{44,394} \\ \hline
\end{tabular}
\caption{Frequency of the four entity types in Wojood and \wojoodf.}
\end{table}

\subsection{subtypes}
All \tagg{GPE, ORG, LOC} and \tagg{FAC} tagged tokens in \wojoodf corpus were annotated with the appropriate subtype based on the context, adding an additional $31$ entity subtypes to \wojoodf. Throughout our annotation process, The \href{https://www.ldc.upenn.edu/sites/www.ldc.upenn.edu/files/arabic-entities-guidelines-v7.4.2.pdf}{ LDC's ACE 2008 annotation guidelines for Arabic Entities V7.4.2} served as the basis for defining our annotation guidelines. Nevertheless, we added new tags (NEIGHBORHOOD, CAMP, SPORT, and ORG\_FAC)  to cover additional cases. Table \ref{tab:subtype_count} lists the frequency of each subtype in \wojoodf. Tables \ref{tab:tab5} and \ref{tab:tab55} in Appendix \ref{app:subtypes} present a brief explanation and examples of each subtype. 

\begin{table}[h!]
\scriptsize

    \begin{tabular}{p{1.0cm}|p{3.5cm}|p{1.5cm}}
\hline
  \textbf{Tag} &  \textbf{Sub-type Tag} & \textbf{Count} \\ \hline
            \hline
            \multirow{7}{*}{GPE} 
                & COUNTRY & 8,205         \\
                & STATE-OR-PROVINCE &  1,890 \\
                & TOWN  &  12,014       \\
                & NEIGHBORHOOD & 119          \\
                & CAMP  &  838                 \\
                & GPE\_ORG  &  1,530              \\
                & SPORT  &  8                     \\
                \hline
         \multirow{9}{*}{LOC} 
                & CONTINENT & 214 \\
                & CLUSTER  &  303 \\
                & ADDRESS  &  0 \\
                & BOUNDARY &  22 \\
                & CELESTIAL  &  4  \\
                & WATER-BODY  &  123 \\
                & LAND-REGION-NATURAL  &  259  \\
                & REGION-GENERAL  & 383  \\
                & REGION-INTERNATIONAL  &  110  \\
                \hline
          \multirow{10}{*}{ORG} 
                & GOV &  8,325  \\
                & COM  &   611  \\
                & EDU  &   1,159      \\
                & ENT &     3     \\
                & NONGOV  &   5,779       \\
                & MED  &    4,111      \\
                & REL  &     96    \\
                & SCI  &     146  \\
                & SPO &      21   \\
                & ORG\_FAC &  114       \\
                \hline
         \multirow{5}{*}{FAC} 
                & PLANT &   1      \\
                & AIRPORT  &  6        \\
                & BUILDING-OR-GROUNDS  &  1017   \\
                & SUBAREA-FACILITY &    134     \\
                & PATH  &  76
                                  \\\hline
                \textbf{Total} & & \textbf{47,621} \\ \hline
 \end{tabular}
\caption{Counts of each subtype entity in the corpus.\label{tab:tab3} }
\label{tab:subtype_count}
\end{table}

\subsection{\wojoodf Annotation Guideline} 
\label{sec:anno_guidelines}

We followed ACE annotation guidelines to annotate the subtypes in \wojoodf. However, since \wojoodf is based on Wojood, we found a discrepancy between Wojood and ACE guidelines. To address this issue in \wojoodf, we reviewed the annotations related to \tagg{GPE, ORG, LOC} and \tagg{FAC} to ensure compatibility with ACE guidelines. In this section, we highlight a number of the challenging annotation decisions we made in \wojoodf.\\

\begin{figure}[h!]  
    \centering       
    \includegraphics[width=0.5\textwidth] {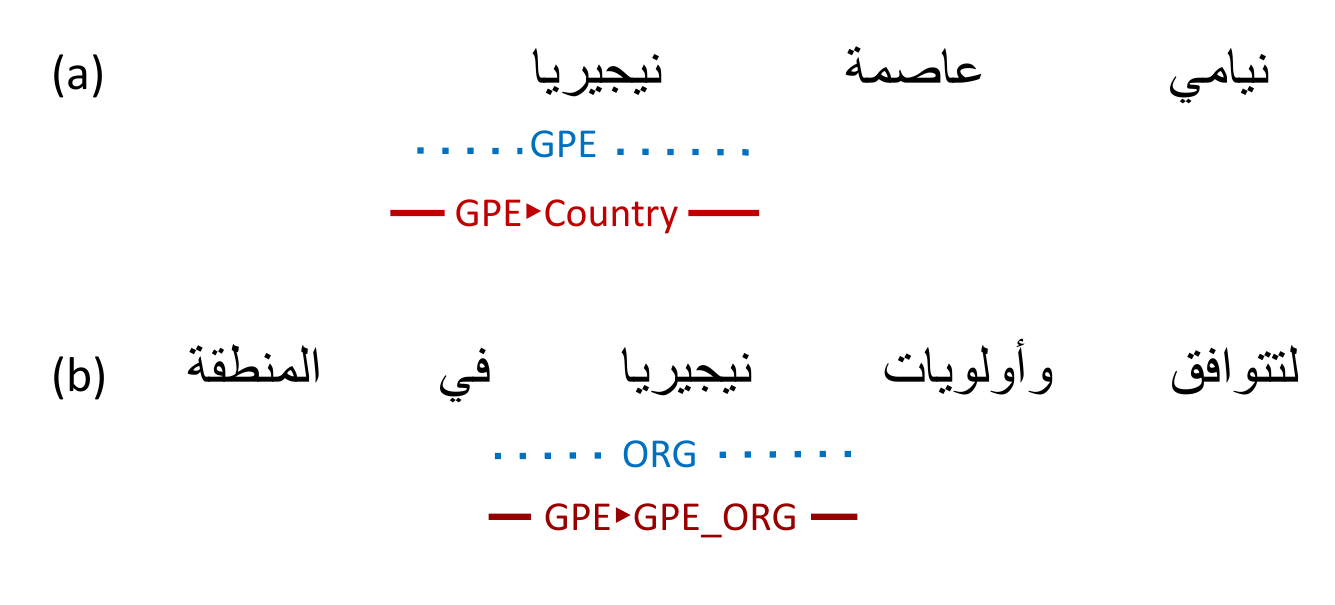}
   \caption{Two examples illustrating the difference between Wojood (in blue) and \wojoodf guidelines (in red) for annotating GPEs.}   
\label{fig0}
\end{figure}

\noindent \textbf{Country's governing body}: in Wojood, country mentions were annotated as  \tagg{GPE} and if the intended meaning of the country is a governing body then it is annotated as  \tagg{ORG}. However, in \wojoodf, all  \tagg{ORG} mentions that refer to the country's governing body are annotated as  \tagg{GPE} with the subtype  \tagg{GPE\_ORG}. 
Figure \ref{fig0} illustrates two examples to illustrate the difference between Wojood and \wojoodf guidelines. According to Wojood, \Ar{نيجيريا} /Nigeria is tagged once as  \tagg{GPE} and once as  \tagg{ORG}, while in \wojoodf both are  \tagg{GPE} in the first level and in the second level one is tagged as \tagg{Country} and the other as \tagg{GPE\_ORG}.

\begin{figure}[h!]  
    \centering       
    \includegraphics[width=0.5\textwidth]{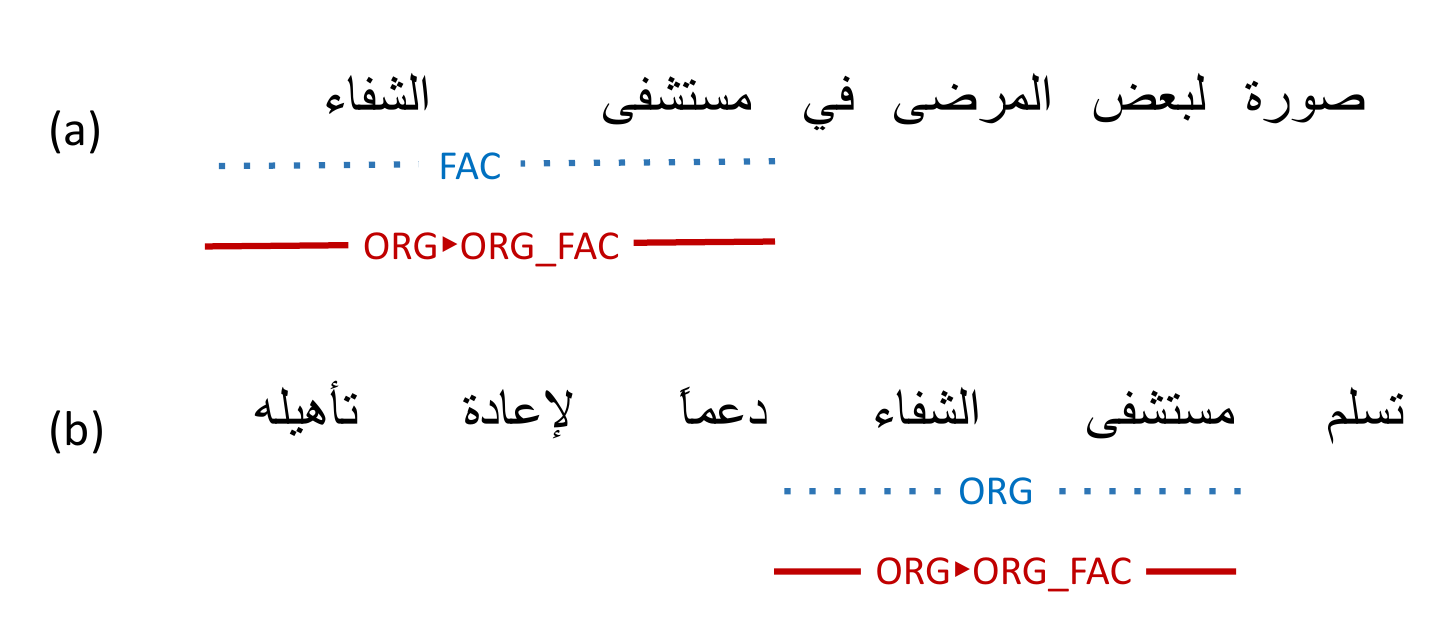}
    \caption{Two examples illustrating the difference between Wojood (in blue) and \wojoodf (in red) guideline for annotating  \tagg{FAC} vs. \tagg{ORG}.}
\label{fig1} 
\end{figure}

\noindent \textbf{Facility vs. organization}: Wojood annotates buildings as  \tagg{FAC} but if the intended meaning, in the context is an organization, then it is annotated as \tagg{ORG}. In \wojoodf, all mentions that refer to the facility's organization or social entity are annotated as \tagg{ORG} with the subtype  \tagg{ORG\_FAC}. Figure \ref{fig1} illustrates an example of this case. Instead of annotating (\Ar{مستشفى الشفاء} /Al-Shifa Hospital) once as  \tagg{FAC} and once as  \tagg{ORG}, \wojoodf tags it as  \tagg{ORG} in the first level, and \tagg{ORG\_FAC} in the second level.

\begin{figure}[h!]
    \centering
     \includegraphics[width=0.5\textwidth]{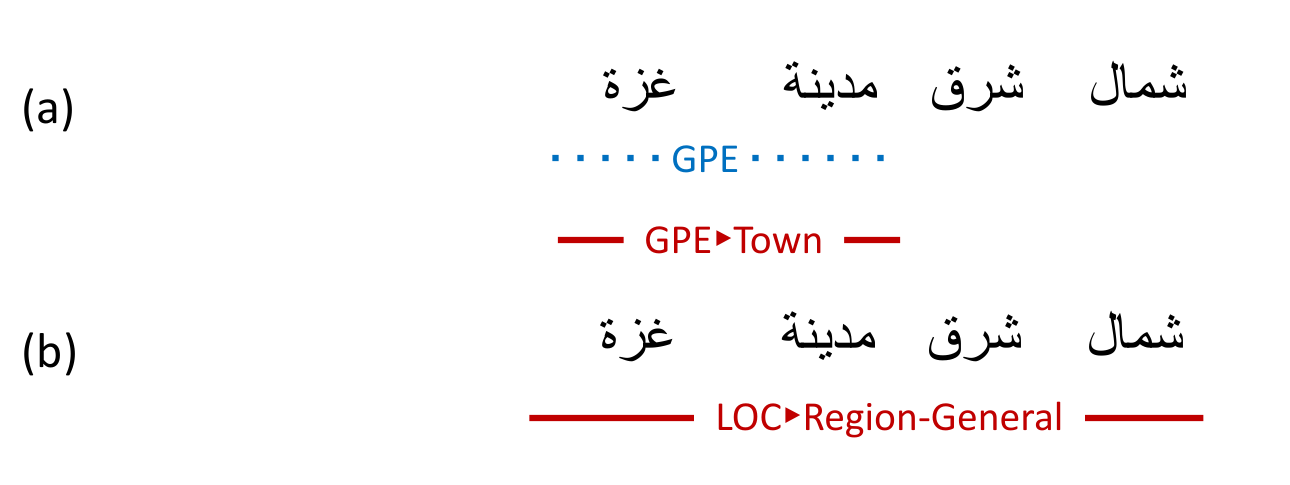}
\caption{(a) The direction (\Ar{شمال شرق مدينة غزة} / north east Gaza city) is not annotated in Wojood, while in (b) it is annotated as \tagg{LOC} with Region-General as subtype in \wojoodf.}
\label{fig2}
\end{figure}

\noindent \textbf{Directions}: Wojood does not include annotations for directions (east, west, south, and north). However, in \wojoodf direction mentions are annotated as  \tagg{LOC} with two subtypes:  \tagg{REGION-GENERAL} if the location does not cross national borders, or  \tagg{REGION-INTERNATIONAL} if the location crosses national borders. See the example in Figure \ref{fig2}. 

In addition to the changes mentioned in this section, ACE guidelines considered any unit that is smaller-size than a village, like neighborhoods or camps, as  \tagg{LOC}, while it is considered as   \tagg{GPE} in Wojood guidelines. Continents are labaled as  \tagg{LOC} in Wojood, while it is \tagg{GPE} in ACE. Both of these cases where corrected in \wojoodf.


\subsection{Annotation Process}
The annotation process was done by one annotator, managed by NER expert, and was conducted over two phases: 

\noindent \textbf{Phase I:} manually revise all annotations of  \tagg{GPE, ORG, LOC,} and  \tagg{FAC} in Wojood according to ACE guidelines, as discussed in section \ref{sec:anno_guidelines}. Table \ref{tab:tab2} shows the counts of each of the four entity types in Wojood
and \wojoodf.\label{tab:tab2}\\

\noindent \textbf{Phase II:} 
manually annotate the  \tagg{GPE, ORG, LOC,} and  \tagg{FAC} with subtypes. The annotator meticulously read each token in every sentence and classified the tokens into their respective subtypes. All critical and problematic tokens are reviewed by the NER expert.\\

\noindent \textbf{Phase III:}
The NER expert reviewed all annotations marked in Phase I and Phase II in order to validate the entities that have been annotated.

Table \ref{tab:tab3} presents the counts of each entity subtype in the corpus, which shows 47,621 annotated entities in total.
 
 \subsection{Inter-Annotator Agreement}
It has been shown that inter-annotator consistency significantly affects the quality of training data and, consequently, a NER system's ability to learn \cite{Zhang2013NamedER}. 
To measure the subtypes annotation quality and consistency, we recruited a second annotator to re-annotate 25,490 tokens (5.0\% of the corpus) that were previously annotated by the first annotator. The sentences were selected randomly from the corpus while diversifying the sources and domains they were selected from. We then assessed the data quality and annotation consistency using the inter-annotator agreement (IAA), measured using Cohen's Kappa ($\kappa$) and $F_1$. The overall IAA was measured at $\kappa=0.9861$ and $F_1=0.9889$. Refer to Table \ref{tab:tab4} for the IAA for each subtype.

One can clearly observe that $\kappa$ is high and that is for multiple reasons. First, we revised the  annotations of the main four entity types \tagg{(GPE, ORG, LOC} and \tagg{FAC)} to better match ACE guideline. Second, once we verified the top level entity types, we started annotating the subtypes. Since the types and subtypes are hierarchically organized, that constraint the number of possible subtypes per token, leading to high IAA. 
Third, the NER expert gave a continuous feedback to the annotator and challenging entity mentions were discussed with the greater team.


As mentioned above, we calculated the IAA using both, Cohen's Kappa and $F_1$, for the subtypes of \tagg{GPE, ORG, LOC} and \tagg{FAC} tags. 
In what follows we explain Cohen's Kappa and $F_1$. Note that $F_1$ is not normally used for IAA, but it is an additional validation of the annotation quality.

\subsubsection{Cohen's Kappa}
To calculate Kappa for a given tag, we count the number of agreements and disagreements between annotators for a given subtype (such as \tagg{GPE\_COUNTRY)}. At the token level, agreements are counted as pairwise matches; thus, disagreements happen when a token is annotated by one annotator (e.g., as \tagg{GPE\_COUNTRY}) and (e.g., as \tagg{GPE\_STATE-OR-PROVINCE}) by another annotator. As such, Kappa is calculated by equation \ref{eq:kappa} \citep{10.1162/089120104773633402}.

\begin{equation}
\label{eq:kappa}
\kappa = \frac{P_o - P_e}{1 - P_e}
\end{equation}

\noindent where $P_o$ represents the observed agreement between annotators and $P_e$ represents the expected agreement, which is given by equation \ref{eq:pe}.

\begin{equation}
\label{eq:pe}
P_e = \frac{1}{N^2}\sum_{T}{n_{T1} \times n_{T2}}
\end{equation}

\noindent where $n_{Ti}$ is the number of tokens labeled with tag $T$ by the $i$th annotator and $N$ is the total number of annotated tokens.

\subsubsection{F-Measure} 
For a given tag $T$, the $F_1$ is calculated according to equation \ref{eq:f1}. We only counted the tokens that at least one of the annotators had labeled with the $T$. We then conducted a pair-wise comparison. $TP$ represents the true positives which is the number of agreements between annotators (i.e. number of tokens labeled \tagg{GPE\_TOWN} by both annotators). If the first annotator disagrees with the second, it is counted as false negatives ($FN$), and if the second disagrees with the first, it is counted as false positives ($FP$), with a total of disagreement being $FN + FP$.

\begin{equation}
\label{eq:f1}
F_1 = \frac{2TP}{2TP+FN+FP}
\end{equation}

\begin{table}[h!]
\small
    \begin{tabular}{p{3.2cm}|p{1.5cm}|p{1.9cm}}
\hline
 \textbf{Sub-Type Tag} &  \textbf{Kappa}  & \textbf {F1-Score}  \\ \hline  \hline
                 COUNTRY &  0.9907 &   00.99
                \\ 
                 STATE-OR-PRONIVCE  &  0.9846 &   00.98
                \\ 
                 TOWN  &  0.9983 &   01.00
                \\ 
                 NEIGHBORHOOD  &  01.00 &   01.00
                \\ 
                 CAMP   &  01.00 &   01.00
                \\ 
                 GPE\_ORG  &  0.9810 &   00.98
                \\ 
                 SPORT  &  01.00 &  01.00
                \\ 
                 CONTINENT  &  01.00 &   01.00
                \\ 
                 CLUSTER   &  0.9589 &   00.96
                \\ 
                 ADDRESS   &  -  &  -
                \\ 
                 BOUNDARY  &  01.00 &   01.00
                \\ 
                 CELESTIAL    &  -  &  -
                \\ 
                 WATER-BODY   &  01.00 &   01.00
                \\ 
                 LAND-REGION-NATURAL  &  0.9333 &   00.93
                \\ 
                 REGION-GENERAL  &  0.9589 &   00.96
                \\ 
                 REGION-INTERNATIONAL  &  0.9231 &   00.92
                \\ 

                 GOV &  0.9760 &  00.98
                \\ 
                 COM  &  01.00 &  01.00
                \\ 
                 EDU   &  0.9807 &  00.98
                \\ 
                 ENT   &  -  &  -
                \\ 
                 NONGOV  &  0.9892 &    00.99
                \\ 
                 MED   &  01.00 &   01.00
                \\ 
                 REL   &  0.9630 &   00.96
                \\ 
                 SCI   &  01.00 &  00.10
                \\ 
                 SPO  &  01.00 &   01.00
                \\ 
                 ORG\_FAC  &  01.00 &   01.00
                \\ 
 
                 PLANT  &  - &  -
                \\ 
                 AIRPORT   &  -  &  -
                \\ 
                 BUILDING-OR-GROUNDS  &  01.00 &  01.00  
                \\ 
                 SUBAREA-FACILITY &  01.00 &  01.00
                \\
                 PATH  &  01.00  &  00.00
\\ 
  \hline
 \textbf{Overall} & \textbf{0.9861} &  \textbf{0.9889}\\ \hline
\end{tabular}
\caption{Overall Kappa and F1-score for each sub-type.\label{tab:tab4} }
\end{table}

\section{Fine-Grained NER Modeling}
\label{sec:modelingNER}
\begin{figure}
\centering
\includegraphics[width=\columnwidth]{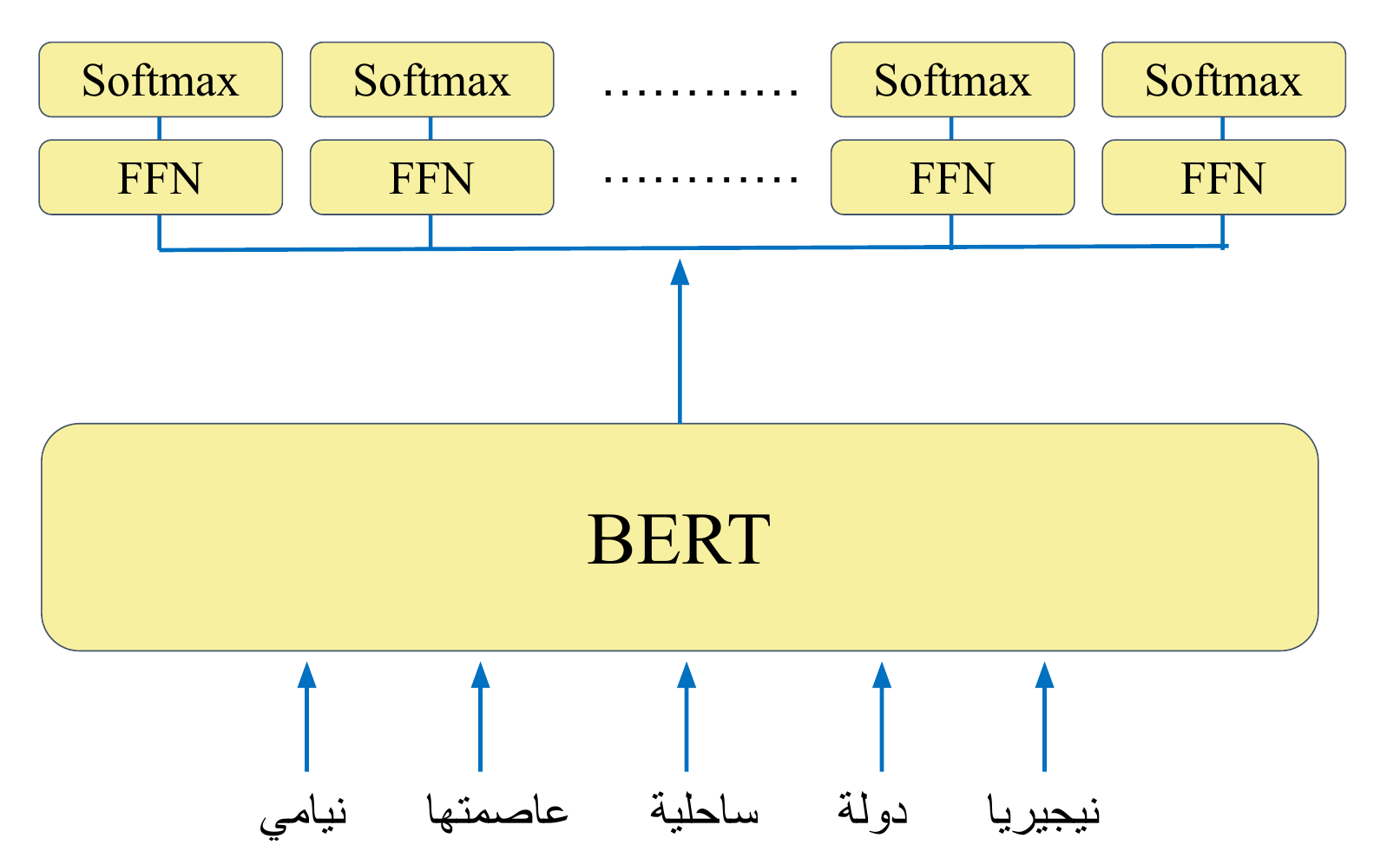}
\caption{ BERT refers to one of three pre-trained models we are using. For flat task, each softmax produce one class for each token, for other tasks each softmax is a set of softmax that produce multiple labels for each token.
}
\label{fig:model_archi}
\end{figure}
\subsection{Approach}
For modeling, we have three tasks all performed on \wojoodf: \textbf{(1)} \textit{Flat NER}, where for each token, we predict a single label from a set of $21$ labels, \textbf{(2)} \textit{Nested NER}, where we predict multiple labels picked from the $21$ tags (i.e., multi-label classification) for each token and \textbf{(3)} \textit{Nested with Subtypes NER}, this is also a multi-label task, where we ask the model to predict the main entity types and subtypes for each token from $52$ total labels. We frame this as multi-task approach since we are learning both the nested labels \textit{and} their subtypes jointly. In the multi-task case, each entity/subtype has its own classification layer, in the case of nested NER and nested with subtypes NER, the model consists of $21$ and $52$ classification layers, respectively. Since we use the IOB2 \cite{IOB2} tagging scheme, each linear layer is a multi-class classifier that outputs the probability distribution through softmax activation function for three classes, $C \in \{I, O, B\}$ \cite{JKG22}. The model is trained with cross entropy loss objective computed for each linear layer separately, which are summed to compute the final cross entropy loss. All models are flat in the sense that we do not use any hierarchical architectures. However, future work can consider employing a hierarchical architecture where nested tokens are learnt first \textit{then} their subtypes within the model. For all tasks, we fine-tune three encoder-based models for Arabic language understanding. Namely, we use ARBERTv2 and MARBERTv2~\citep{elmadany2023orca}, which are both improved versions of ARBERT and MARBERT~\citep{abdul-mageed-etal-2021-arbert}, respectively, that are trained on bigger datasets. The third model is ARABERTv2, which is an improved version of ARABERT~\cite{antoun2021arabert}. It is also trained on a bigger dataset, with improved preprocessing. Figure~\ref{fig:model_archi} offers a simple visualization of our models' architecture.

\subsection{Training Configuration}
\label{Training_config}
\newcommand{\plustopminus}{\stackon[0.1pt]{$-$}{$\scriptscriptstyle+$}}

\begin{table}
    \centering
    \setlength{\tabcolsep}{3.5pt}
    \renewcommand{\arraystretch}{0.5}
    \begin{tabular}{l|l|l|l}
        \hline
        Task & Model & Dev & Test \\
        \hline
        \multirow{3}{*}{Flat} & M1 & 0.917\textsuperscript{±0.00} & \textbf{0.920}\textsuperscript{±0.00} \\
                              & M2 & 0.910\textsuperscript{±0.00} & 0.913\textsuperscript{±0.01} \\
                              & M3 & 0.902\textsuperscript{±0.00} & 0.907\textsuperscript{±0.01} \\ \hline
        \multirow{3}{*}{Nested } & M1 & 0.844\textsuperscript{±0.02} & 0.845\textsuperscript{±0.01} \\
                                & M2 & 0.868\textsuperscript{±0.02} & 0.861\textsuperscript{±0.02} \\
                                & M3 & 0.858\textsuperscript{±0.02} & \textbf{0.866}\textsuperscript{±0.02} \\ \hline
        \multirow{3}{*}{\makecell{Nested\\+subtypes}} & M1 & 0.836\textsuperscript{±0.01} & 0.837\textsuperscript{±0.01} \\
                                 & M2 & 0.880\textsuperscript{±0.01} & 0.883\textsuperscript{±0.01} \\
                                 & M3 & 0.883\textsuperscript{±0.00} & \textbf{0.885}\textsuperscript{±0.00} \\ \hline
    \end{tabular}
    \caption{Results of fine-tuned models on the three different tasks. \textbf{M1}: ARBERTv2, \textbf{M2}: MARBERTv2 and \textbf{M3}: ARABERTv2. The results are represented as F1 averaged over 3 runs.}
    \label{tab:all_tasks_table}
\end{table}

We split our dataset into three distinct parts for training (Train) $70$\%, validation (Dev) $10$\%, and blind testing (Test) $20$\%. We fine-tune all three models for $50$ epochs each with an early stopping patience of $5$ as identified on Dev. We use the AdamW optimizer~\cite{loshchilov2019decoupled}, an exponential learning rate scheduler and a dropout of $0.1$. The maximum sequence length is $512$, the batch size, $B=8$, and the learning rate, $\eta=1e^{-5}$. For each model, we report an average of three runs (each time with a different seed). We report in $F_1$ along with the standard deviation from the three runs, on both Dev and Test, for each model. All models are implemented using PyTorch, Huggingface Transformers, and a custom version of the Wojood open-source code\footnote{\href{https://github.com/SinaLab/ArabicNER}{https://github.com/SinaLab/ArabicNER}}.

\subsection{Results}
We show the results of our three fine-tuned models across each of the three tasks in Table \ref{tab:all_tasks_table}. We briefly highlight these results in the following:

\noindent \textbf{Flat NER.} The three fine-tuned models achieve comparable results on the Flat NER task, with ARBERTv2 scoring slightly better on both the Dev and Test sets. ARBERTv2 achieves an $F_1$ of $92\%$ on the Test set, while ARBERTv2 and ARABERTv2 achieves $91.3\%$ and $90.3\%$, respectively.

\noindent \textbf{Nested NER.} ARABERTv2 slightly outperforms other pre-trained models with a small margin, on Dev and Test. On Test, it scores $86.6\%$. 

\noindent \textbf{Nested NER with Subtypes.} Here, ARABERTv2 achieves the highest score ($88.5\% F_1$). 

\section{Analysis}
\label{sec:analysis}
\begin{figure}[ht]
\centering
\includegraphics[width=\columnwidth]{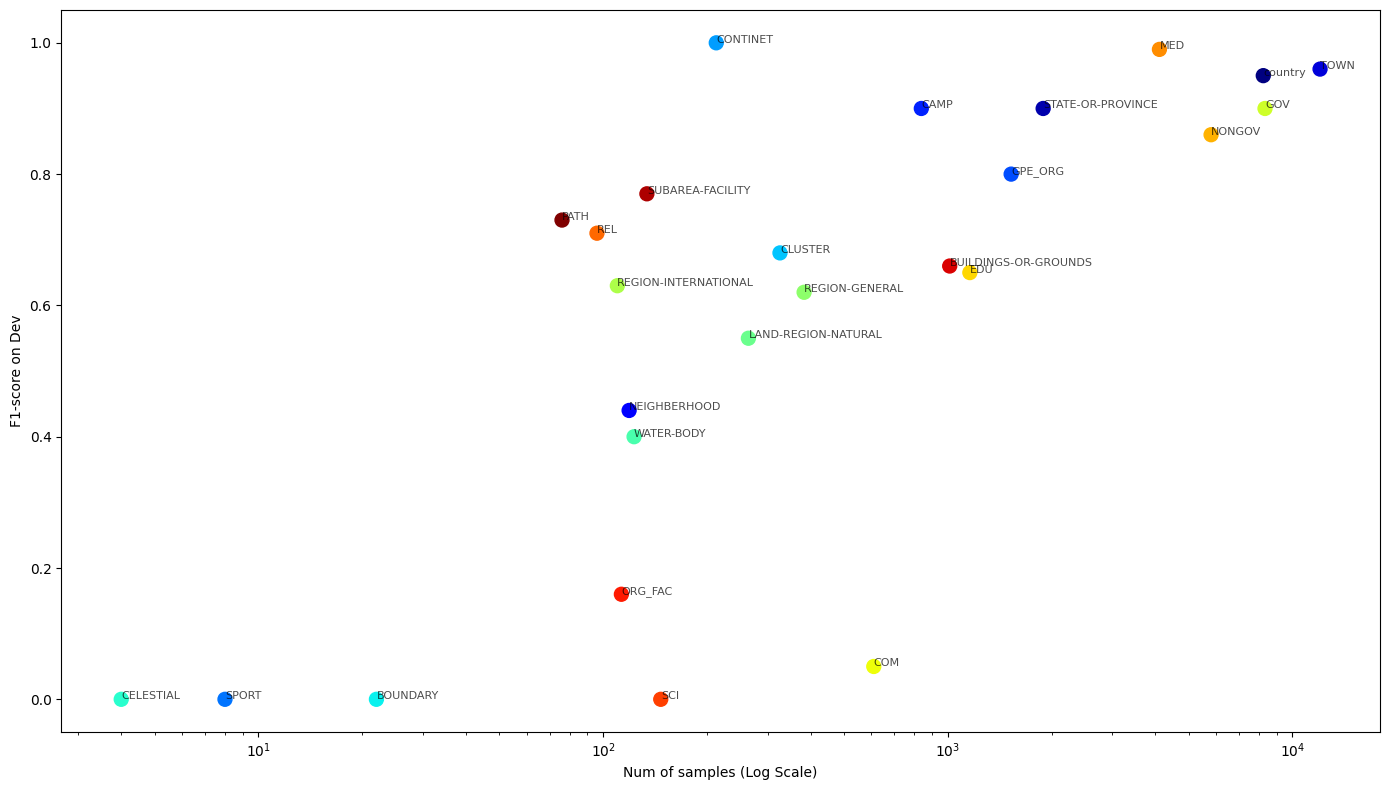}
\caption{ Number of samples vs. $F_1$ in each subtype class on Subtype classification task.}
\label{fig:f1vssamples}
\end{figure}
For all tasks, all models almost always converge in the first $10$ epochs. For all models, there is a positive correlation between performance and the number of training samples. For example, for classes represented well in the training set (e.g., \tagg{COUNTRY}, \tagg{TOWN} and \tagg{GOV}), models perform at 0.90 $F_1$ or above. 

The inverse is also true, with poor performance on classes such as \tagg{SPORT}, \tagg{BOUNDARY} and \tagg{CELESTIAL}. There are also some nuances. For example, we can see that the best model is struggling with the \tagg{COM} subtype class even though the model has scored good results with classes with fewer samples such as \tagg{CLUSTER}. The main reason for this is that types such as \tagg{CLUSTER} are a closed set of classes (e.g., "European Union", "African Union") where the model can easily memorize them, while the \tagg{COM} refers to an infinite group of commercial entities, that can not be limited. Figure~\ref{fig:f1vssamples} is a plot of the number of samples in training data (X-axis) vs. performance (Y-axis) that clearly shows the general pattern of good performance positively correlating with the number of training samples.  

\subsection{Out-of-Domain Performance} 

\begin{figure}[h!]  
    \centering       
    \includegraphics[width=0.5\textwidth]{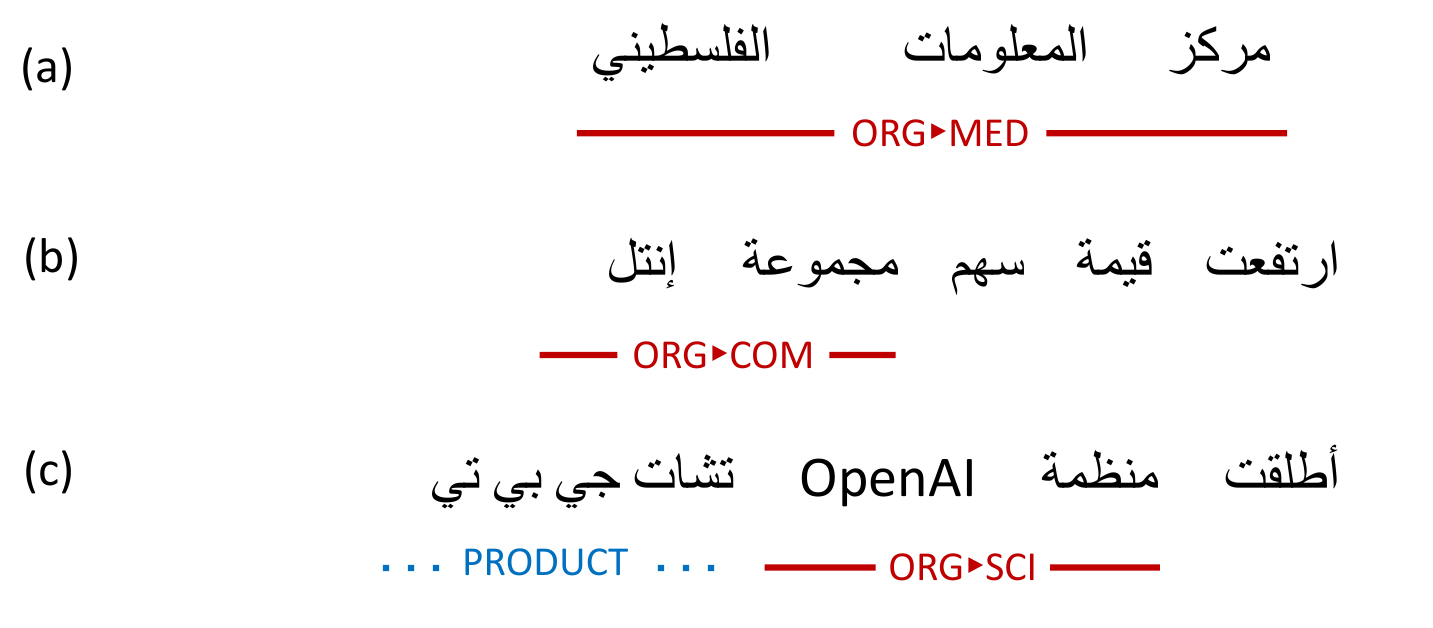}
   \caption{Some mentions from the three new domains that have not previously appeared in \wojoodf. (a) (\Ar{مركز المعلومات الفلسطيني})  in Politics domain, (b) (\Ar{مجموعة إنتل}) in Finance domain, (c) (\Ar{منظمة } OpenAI)  \normalsize in Science domain.}
\label{fig3}
\end{figure}

To assess the generalization capability of our models, we conducted an evaluation on three unseen domains and different time periods. Three corpora were collected, each covering a distinct domain: finance, science, and politics. These corpora were compiled from Aljazeera news articles published in 2023. Manual annotation of the three corpora was performed in accordance with the same annotation guidelines established for \wojoodf. We apply the three versions of each of our three models trained on \wojoodf original training data (described in Section~\ref{Training_config}) on the new domains, for each of the three NER tasks. We present results for this out-of-domain set of experiments in Table~\ref{tab:new_domains_table}. We observe that performance drastically drops on all three new domains, for all models on all tasks. This is not surprising, as challenges related to domain generalization are well-known in the literature. Our results here, however, allow us to quantify the extent to which model performance degrades on each of these three new domains. In particular, models do much better on the politics domain than they perform on finance or science. This is the case since our training data are collected from online articles involving news and much less content from financial or scientific sources. 
Figure \ref{fig3} shows some examples for new mentions from those domains that have not been seen in \wojoodf.

\begin{table}[]
\centering
    \footnotesize
    \setlength{\tabcolsep}{4pt}
    \renewcommand{\arraystretch}{0.9}
    \begin{tabular}{l|l|l|l|l}
        \hline
        Task & Model & Finance & Science & Politics \\ \hline
        \multirow{3}{*}{Flat} & M1 & 63.7\% \textsuperscript{±0.01} & 0.670\textsuperscript{±0.02} & \textbf{0.747}\textsuperscript{±0.02} \\
                              & M2 & 0.573\textsuperscript{±0.01} & \textbf{0.677}\textsuperscript{±0.02} & 0.717\textsuperscript{±0.01} \\
                              & M3 & \textbf{0.643}\textsuperscript{±0.01}& 0.670\textsuperscript{±0.02}& 0.723\textsuperscript{±0.01} \\ \hline
        \multirow{3}{*}{Nested} & M1 & 0.458\textsuperscript{±0.01} & 0.494\textsuperscript{±0.02} & 0.557\textsuperscript{±0.00} \\
                               & M2 & 0.499\textsuperscript{±0.05} & 0.554\textsuperscript{±0.00} & 0.612\textsuperscript{±0.01} \\
                               & M3 & \textbf{0.563}\textsuperscript{±0.02} & \textbf{0.583}\textsuperscript{±0.02} & \textbf{0.629}\textsuperscript{±0.03} \\ \hline
        \multirow{3}{*}{\makecell{Nested\\+subtypes}} & M1 & 0.449\textsuperscript{±0.07} & 0.493\textsuperscript{±0.02} & 0.497\textsuperscript{±0.01} \\
                              & M2 & 0.504\textsuperscript{±0.03} & 0.544\textsuperscript{±0.06} & 0.575\textsuperscript{±0.02} \\
                              & M3 & \textbf{0.553}\textsuperscript{±0.04} & \textbf{0.545}\textsuperscript{±0.02} & \textbf{0.593}\textsuperscript{±0.08} \\ \hline
    \end{tabular}
    \caption{Results of fine-tuned models on the three new domains, Finance, Science, and Politics. \textbf{M1}: MARBERTv2, \textbf{M2}: ARBERTv2 and \textbf{M3}: ARABERTv2. The results are represented as F1 averaged over 3 runs.}
    \label{tab:new_domains_table}
\end{table}

\subsection{Error Analysis}

\begin{table*}
\centering
\setlength{\tabcolsep}{1pt}
\small
\begin{tabular}{p{8.3cm}p{2.4cm}p{2.4cm}p{2.8cm}}\hline
Example                                        & Gold        & Predicted & Error Type          \\ \hline
\Ar{أنا ازا هاجرت ع أي مكان رح اَخد الشلة } & O & GPE|TWN & msa\_dia\_confusion \\
If I ever migrated somewhere, I'd take the group & & & \\ 

\Ar{مشهد 3 فتاة جالسة و خلفها العلم الأمريكي. } & CRDNAL & ORDNAL & ordinal\_vs\_cardinal \\
Scene 3: a girl sitting with the American flag behind her. & & & \\ 

\Ar{ جدار الفصل العنصري مستعمرة بزغات زئيف. } & LOC|NEIGHB & NEIGHB & Missing\_parent\_tag \\
The racial separation wall, colony of Bazgat Ze'ev. & & & \\ 

\Ar{ بتنتخب رئيس جمهورية و رئيس مجلس نواب } & OCC|ORG|GOV & OCC|ORG & missing\_subtype \\
The president of the republic and the speaker of the council of deputies are elected. & & & \\ 

\Ar{ صحيح الساعة خمسة حسب اعلانهم } & TIME & CRDNL & wrong\_tag \\
It's true, it's five o'clock according to their announcement. & & & \\ 

\Ar{ العلما اللغة التانية بتنحصر للاستخدام اليومي. } & B-ORDNL & O & no\_prediction \\
Scientists: the second language is limited to daily use. & & & \\ \hline

\end{tabular}
\caption{Examples of error categories made by our best model (ARABERTv2) on our Dev set. We provide the translation to English of each sample. }
\label{tab:error_examples}
\end{table*}

\begin{figure}[ht]
\centering
\includegraphics[width=\columnwidth]{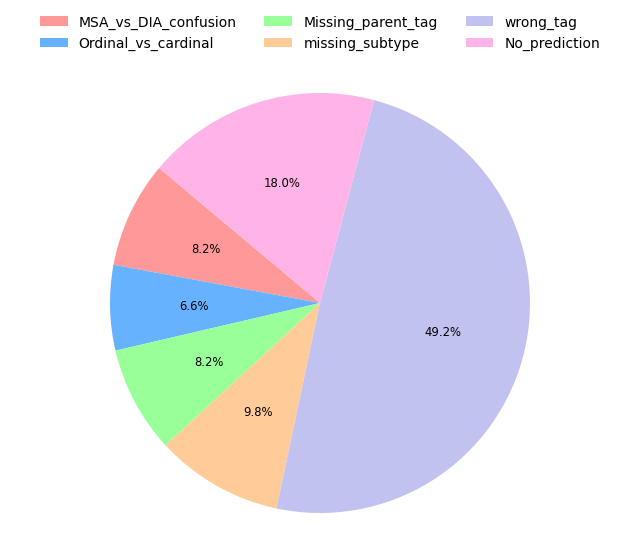}
\caption{Distribution of error classes in nested with subtypes task on our Dev set.}
\label{fig:ErrorsDistribution}
\end{figure}

In order to understand the errors made by the model, we conduct a human error analysis on the errors generated by ARABERTv2 (i.e, best model on this task) on the first $2$K tokens of the Dev set of Nested NER with Subtypes task. We find that the model's errors can be categorized into six major error classes: \textbf{(1)} \textit{wrong tag}, where the model predicts a different tag, \textbf{(2)} \textit{no prediction}, where the model does not produce any tag (i.e. predict O), \textbf{(3)} \textit{missing subtype}, the model succeeds in predicting parent tag but fails to predict the subtype, \textbf{(4)} \textit{missing parent tag}: the model succeeds in predicting subtype tag but fails to predict the parent tag, \textbf{(5)} \textit{MSA vs. DIA confusion}, the model makes a wrong prediction due to confusion between MSA and Dialect, and \textbf{(6)} \textit{ordinal vs. cardinal}, in this class, the model assigns cardinal to an ordinal class. Figure~\ref{fig:ErrorsDistribution} shows the distribution of different errors present in the Dev set, with the \textit{wrong tag} being the major source of errors followed by \textit{no prediction} error. A further breakdown of the \textit{wrong tag} error class shows that $14.3$\% are due to usage of dialectal words, a similar proportion are due to nested entities. Table \ref{tab:error_examples} shows an example of each error class.

\section{Conclusion and Future Work}
\label{sec:conclusion}
We presented \wojoodf, an extension to the Wojood NER corpus with subtypes for the \tagg{GPE}, \tagg{LOC}, \tagg{ORG}, and \tagg{FAC}. \wojoodf corpus is the first fine-grain corpus for MSA and dialectal Arabic with nested and subtyped NER. The \tagg{GPE}, \tagg{ORG}, \tagg{FAC} and \tagg{LOC} tags form more than $44$K tokens of the corpus, which was manually annotated using subtypes entities. 
Our inter-annotator agreement IAA evaluation of \wojoodf annotations achieved high levels of agreement among the annotators. The achieved evaluations are 0.9861 Kappa and 0.9889 $F_1$. 

We also fine-tune three pre-trained models ARBERTv2, MARBERTv2 and ARABERTv2 and tested their performance on different settings of \wojoodf. We find that ARABERTv2 achieved the best performance on Nested and Nested with Subtypes tasks. In the future, we plan to test pre-trained models on nested subtypes with hierarchical architecture. We also plan to link named entities with concepts
in the Arabic Ontology \cite{J21,J11} to
enable a richer semantic understanding of text. Additionally, we will extend the \wojoodf corpus to include more dialects, especially the Syrian Nabra dialects \cite{ANMFTM23} as well as the four dialects in the Lisan \cite{JZHNW23} corpus.

\section*{Acknowledgment}
\label{sec:ack}
We would like to thank  Sana Ghanem for her invaluable assistance in reviewing and improving the annotations, as well as for her support in the IAA calculations. The authors would also like to thank Tymaa Hammouda for her technical support and expertise in the data engineering of the corpus.

\section*{Limitations}
A number of considerations related to limitations and ethics are relevant to our work, as follows:
\begin{itemize}
    \item \textbf{Intended Use.} Our models perform named entity recognition at a fine-grained level and can be used for a wide range of information extraction tasks. As we have shown, however, even though the models are trained with data acquired from several domains, their performance drops on data with distribution different than our training data such as the finance or science domains. We suggest this be taken into account in any application of the models. 
    \item \textbf{Annotation Guidelines and Process.} Some of the entities are difficult to tag. Even though annotators have done their best and we report high inter-annotator reliability, the application of our guidelines may need to be adapted before application to new domains.
\end{itemize} 

\section*{Ethics Statement}
We trained our models on publicly available data, thus we do not have any particular concerns about privacy.

\bibliography{anthology,custom,MyReferences}
\bibliographystyle{acl_natbib}

\appendix
\section{subtypes and Inter-Annotator Agreement}
\label{app:subtypes}
This is an appendix contains \wojoodf subtype descriptions and detailed IAA. 

\begin{table*}[h!]
\centering 
    \begin{tabular}{|p{1.2cm}|p{3.0cm}|p{10.5cm}|}
\hline
  \textbf{Tag} &  \textbf{Sub-type Tag} & \textbf{Short Description} \\ \hline
            \hline
            \multirow{7}{*}{GPE} 
                & COUNTRY & Taggable mentions of the entireties of any nation.  
                \Ar{فلسطين، مصر،} \Ar{الولايات المتحدة، لبنان.}
                                  \\\cline{2-3}
                & STATE-OR-PRONIVCE &  Taggable mentions of the entireties of any state, province, or canton of any nation.
                 \Ar{محافظة القاهرة، قطاع غزة،}
               \Ar{إقليم كردستان، لواء نابلس.}
                                  \\\cline{2-3}
                & TOWN  &  Taggable mentions of any GPE entireties below the level of State-or-Province, including cities, and villages.
                 \Ar{،العاصمة دبي،}
               \Ar{قرية بيرزيت.}
                                  \\\cline{2-3}
                & NEIGHBORHOOD &  Taggable mentions of the entireties of units that are smaller than villages.
                 \Ar{حي الطيرة، البلدة القديمة، حي المغاربة.}
                                  \\\cline{2-3}
                & CAMP  &  Taggable mentions of the entireties of units that are smaller than villages, relating to refugees.
                 \Ar{مخيم قلنديا، مخيم نور شمس.}
                                  \\\cline{2-3}
                & GPE\_ORG  &  is used for GPE mentions that refer to the entire governing body of a GPE.
                \Ar{ أصدرت الولايات المتحدة تقريرها،}
               \Ar{قررت فلسطين إعفاء المتضررين.}
                                  \\\cline{2-3}
                & SPORT  &  Athletes, Sports Teams.
                 \Ar{مباراة المغرب، الفرق الرياضية.}
               \Ar{ برشلونة، ميلان.}
                                  \\\cline{2-3}
                \hline
         \multirow{9}{*}{LOC} 
                & CONTINENT & Taggable mentions of the entireties of any of the seven continents.  
                \Ar{أوروبا، آسيا.}  
                                  \\\cline{2-3}
                & CLUSTER  &  Named groupings of GPEs that can function as political entities.
                \Ar{أوروبا الشرقية، الشرق الأوسط.}
                                  \\\cline{2-3}
                & ADDRESS  &  A location denoted as a point such as in a postal system ("31° S, 22° W").
                 \Ar{17، شارع فؤاد.}
                                  \\\cline{2-3}
                & BOUNDARY &  A one-dimensional location such as a border between GPE’s or other locations. 
                 \Ar{الحدود الشرقية، الحدود السورية التركية.}
                                  \\\cline{2-3}
                & CELESTIAL  &  world, earth, globe in addition to all other planets. 
                 \Ar{ المريخ، عطارد.}
                                  \\\cline{2-3}
                & WATER-BODY  &  Bodies of water, natural or artificial (man-made). 
                \Ar{ البحر الأحمر، الأطلسي.}
                                  \\\cline{2-3}
                & LAND-REGION-NATURAL  &  Geologically or ecosystemically designated, non-artificial locations.
                \Ar {جبال الألب، الأغوار،  السهول.}
                                    \\\cline{2-3}
                & REGION-GENERAL  & Taggable locations that do not cross national borders.
               \Ar  {شمال الضفة الغربية، شرق سوريا.}
                                  \\\cline{2-3}
                & REGION-INTERNATIONAL  &  Taggable locations that cross national borders.
                \Ar {آسيا الكبرى، جنوب أفريقيا.}
                                  \\\cline{2-3}
  \hline
 \end{tabular}
\caption{Parent type and description of each sub-type in \wojoodf \label{tab:tab5} }
\end{table*}
\begin{table*}[h!]
\centering 
     \begin{tabular}{|p{1.2cm}|p{3.0cm}|p{10.5cm}|}
\hline
  \textbf{Tag} &  \textbf{Sub-type Tag} & \textbf{Short Description} \\ \hline
            \hline
          \multirow{10}{*}{ORG} 
                & GOV &  Government organizations.
                \Ar{سفارة، محكمة، وزارة، شرطة .}
                                  \\\cline{2-3}
                & COM  &   A commercial organization that is focused primarily upon providing ideas, products, or services for profit.
                \Ar{بنك ، شركة ،مؤسسة ربحية.}
                                  \\\cline{2-3}
                & EDU  &   An educational organization that is focused primarily upon the furthering or promulgation of learning/education.
                \Ar{جامعة، مدرسة، معهد.}
                                  \\\cline{2-3}
                & ENT &   Entertainment organizations whose primary activity is entertainment.
                \Ar{فرقة ميامي، مسرح الحكواتي.}
                                  \\\cline{2-3}
                & NONGOV  &   Non-governmental organizations that are not a part of a government or commercial organization and whose main role is advocacy, charity or politics (in a broad sense).
                \Ar{نقابة العاملين، الأمم المتحدة، الأحزاب السياسية ،أطباء بلا حدود.}
                                  \\\cline{2-3}
                & MED  &   Media organizations whose primary interest is the distribution of news or publications.
                \Ar{جريدة الشرق، مجلة الحياة.}
                                  \\\cline{2-3}
                & REL  &  Religious organizations that are primarily devoted to issues of religious worship. 
                \Ar{الأوقاف ، الأزهر.}
                                  \\\cline{2-3}
                & SCI  & Medical-Science organizations whose primary activity is the application of medical care or the pursuit of scientific research.
                \Ar{مستشفى هداسا ،معهد الدراسات النووية.}
                                  \\\cline{2-3}
                & SPO &  Sports organizations that are primarily concerned with participating in or governing organized sporting events.
                \Ar{ الاتحاد السعودي لكرة القدم، لجنة الفلبين الأولومبية.}
                                  \\\cline{2-3}
                & ORG\_FAC &  Facilities that have an organizational, legal or social representative
                \Ar{مظاهرات أمام بنك روما.}
                                  \\\cline{2-3}
                \hline
         \multirow{5}{*}{FAC} 
                & PLANT &  One or more buildings that are used and/or designed solely for industrial purposes: manufacturing, power generation, etc.
                \Ar{مصنع.}
                                  \\\cline{2-3}
                & AIRPORT  &   A facility whose primary use is as an airport.
                \Ar{مطار.}
                                  \\\cline{2-3}
                & BUILDING-OR-GROUNDS  &   Man-made/-maintained buildings, outdoor spaces, and other such facilities. 
                \Ar{ منزل، مبنى، مستشفى، معبر.}
                                  \\\cline{2-3}
                & SUBAREA-FACILITY &  Taggable portions of facilities.
                \Ar{غرفة ،زنزانة.}
                                  \\\cline{2-3}
                & PATH  &  Streets, canals, and bridges.
                \Ar{ الشوارع الرئيسية، الخطوط الهاتفية، الحواجز.}
                                  \\\hline
 \end{tabular}
\caption{Parent type and description of each sub-type in \wojoodf \label{tab:tab55} }
\end{table*}
\begin{table*}[h!]
  \centering
\begin{tabular}{| *{6}{c} |}
\hline
  \textbf{Sub-type Tag} & \textbf{TP}  & \textbf{FN} & \textbf{FP}  &  \textbf{Kappa}  & \textbf {F1-Score}  \\ \hline  \hline
                 COUNTRY & 643 &  5 &  7 &  0.9907 &   00.99
                \\ 
                 STATE-OR-PRONIVCE & 96 &  3 &  0 &  0.9846 &  00.98
                \\ 
                 TOWN  & 295 &  0 &  1 &  0.9983 &   01.00
                \\ 
                 NEIGHBORHOOD & 23 &  0 &  0 &   01.00 &  01.00
                \\ 
                 CAMP  & 92 &  0 &  0 &  01.00 &  01.00
                \\ 
                 GPE\_ORG  & 129 &  3 &  2 &  0.9810 &   00.98
                \\ 
                 SPORT  & 2 &  0 &  0 &  01.00 &  01.00
                \\ 
                 CONTINENT & 7 &  0 &  0 &  01.00 &   01.00
                \\ 
                 CLUSTER  & 35 &  3 &  0 &  0.9589 &  00.96
                \\ 
                 ADDRESS  &  - &  - &  - &  -  &  -
                \\ 
                 BOUNDARY & 11 &  0 &  0 &  01.00 &   01.00
                \\ 
                 CELESTIAL   &  - &  - &  - &  - &  -
                \\ 
                 WATER-BODY  & 5 &  0 &  0 &  01.00 &   01.00
                \\ 
                 LAND-REGION-NATURAL  & 14 &  0 &  2 &  0.9333 &   00.93
                \\ 
                 REGION-GENERAL  & 70 &  2 &  4 &  0.9589 &  00.96
                \\ 
                 REGION-INTERNATIONAL  & 6 &  0 &  1 &  0.9231 &   00.92
                \\ 

                 GOV & 490 &  6 &  18 &  0.9760 &  00.98
                \\ 
                 COM  & 21 &  0 &  0 &  01.00 &   01.00
                \\ 
                 EDU  & 153 &  0 &  6 &  0.9807 &   00.98
                \\ 
                 ENT  &  - &  - &  - &  - &  -
                \\ 
                 NONGOV  & 599 &  11 &  2 &  0.9892 &   00.99
                \\ 
                 MED  & 630 &  0 &  0 &  01.00  &  01.00
                \\ 
                 REL  & 26 &  2 &  0 &  0.9630 &  00.96
                \\ 
                 SCI  & 4 &  0 &  0 &  01.00 &   00.10
                \\ 
                 SPO & 2 &  0 &  0 &  01.00 &  01.00
                \\ 
                 ORG\_FAC & 15 &  0 &  0 &  01.00 &   01.00
                \\ 
 
                 PLANT & - &  - &  - &  - &   -
                \\ 
                 AIRPORT   &  - &  - &  - &  - &  -
                \\ 
                 BUILDING-OR-GROUNDS  & 64 &  0 &  0 &  01.00 &   01.00  
                \\ 
                 SUBAREA-FACILITY & 48 &  0 &  0 &  01.00 &   01.00
                \\
                 PATH  & 2 &  0 &  0 &  01.00 &   01.00
\\ 
  \hline
 \textbf{Overall} & \textbf{3,482 \scriptsize count}  & \textbf{35 \scriptsize count} & \textbf{43 \scriptsize count} & \textbf{0.9861 \scriptsize macro} & \textbf{0.9889 \scriptsize micro}\\ \hline
\end{tabular}
\caption{Overall IAA for each sub-type, reported using Kappa and $F_1$. 
\label{tab:tab10} }
\end{table*} 

\end{document}